\documentclass{article}
\usepackage[utf8]{inputenc}
\usepackage[numbers]{natbib}
\usepackage{multirow,makecell}
\usepackage{xcolor}
\usepackage{graphicx}
\usepackage{longtable}
\usepackage[colorinlistoftodos]{todonotes}
\newcommand{\tasksum}{\textit{sum digits} }
\newcommand{\taskhowmany}{\textit{how many 3 or 4} }

    \usepackage[preprint]{neurips_2023}

\usepackage[utf8]{inputenc} %
\usepackage[T1]{fontenc}    %
\usepackage{hyperref}       %
\usepackage{url}            %
\usepackage{booktabs}       %
\usepackage{amsfonts}       %
\usepackage{nicefrac}       %
\usepackage{microtype}      %
\usepackage{xcolor}         %

\begin{document}

\title{On the Promise for Assurance of Differentiable Neurosymbolic Reasoning Paradigms}

\author{%
  Luke E. Richards\textsuperscript{1,2}\thanks{Corresponding author: \texttt{firstname.lastname@pnnl.gov}} \\ 
  \And
  Jessie Yaros\textsuperscript{1} \\
 \And
  Jasen Babcock\textsuperscript{1} \\
   \And
   \And
  Coung Ly\textsuperscript{1} \\
   \And
  Robin Cosbey\textsuperscript{1} \\
   \And
  Timothy Doster\textsuperscript{1} \\
   \And
  Cynthia Matuszek\textsuperscript{2}
}

\maketitle

\vspace{-2.5em} %
\begin{center}
\textsuperscript{1}Pacific Northwest National Laboratory (PNNL) \\
\textsuperscript{2}University of Maryland, Baltimore County (UMBC) \\
\end{center}

\begin{abstract}

To create usable and deployable Artificial Intelligence (AI) systems, there requires a level of assurance in performance under many different conditions. Many times, deployed machine learning systems will require more classic logic and reasoning performed through neurosymbolic programs jointly with artificial neural network sensing. While many prior works have examined the assurance of a single component of the system solely with either the neural network alone or entire enterprise systems, very few works have examined the assurance of integrated neurosymbolic systems. Within this work, we assess the assurance of end-to-end fully differentiable neurosymbolic systems that are an emerging method to create data-efficient and more interpretable models. We perform this investigation using Scallop, an end-to-end neurosymbolic library, across classification and reasoning tasks in both the image and audio domains. We assess assurance across adversarial robustness, calibration, user performance parity, and interpretability of solutions for catching misaligned solutions. We find end-to-end neurosymbolic methods present unique opportunities for assurance beyond their data efficiency through our empirical results but not across the board. We find that this class of neurosymbolic models has higher assurance in cases where arithmetic operations are defined and where there is high dimensionality to the input space, where fully neural counterparts struggle to learn robust reasoning operations. We identify the relationship between neurosymbolic models' interpretability to catch shortcuts that later result in increased adversarial vulnerability despite performance parity. Finally, we find that the promise of data efficiency is typically only in the case of class imbalanced reasoning problems.

\end{abstract}

\section{Introduction}

The need to assure the performance and deployment of Artificial Intelligence (AI)-enabled systems is hitting a peak as many demonstrations of the technology have shown applicability to a growing number of tasks in a plethora of domains. Modern methods, especially in Machine Learning (ML) model development, focus on advances in deep learning or neural networks. Such models create many opportunities but also obfuscate our ability to model behavior for testing assurance encompassing safety, security, cross user-performance, and interpretability. When deploying AI to critical applications like controlling industrial systems, autonomous operation, and other applications with risks for people and their environments, stakeholders are rightfully demanding higher levels of assurance. The past decade in particular has outlined many risks of lack of cross user-performance \cite{danks2017algorithmic}, lack of robustness to data drift \cite{taori2020measuring}, and security vulnerabilities \cite{carlini2024remote} for systems using machine learning.

\begin{figure}[h]
    \centering
    \includegraphics[width=\linewidth]{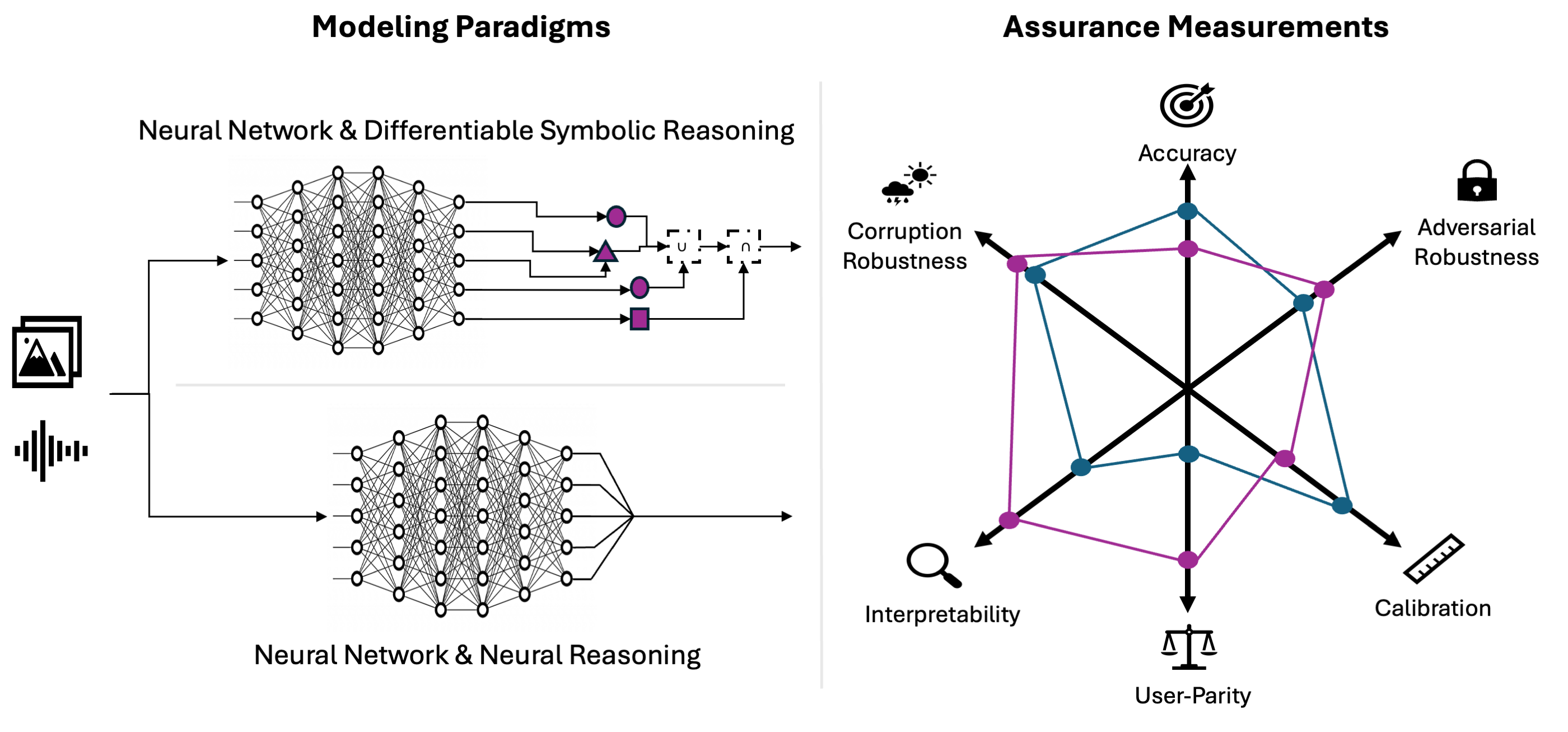}
    \caption{Visualizations of the two paradigms examined and the assurance metrics measured.}
    \label{fig:example}
\end{figure}

The concept of assured-by-design methods attempts to help address this by rethinking the fundamental building blocks of our AI systems to have these factors core to initial design of methods. However, this comes at a time when the basic building blocks in the modern paradigm are being scaled and given more autonomy in interacting with the digital and physical world. Assured-by-design methods can target representational learning \cite{bynum2020rotational}, guarantees or certification of performance under stress \cite{cohen2019certified, rosenfeld2020certified}, model architecture changes \cite{gu2014towards}, and systems-of-systems enclosing networks \cite{zoppi2021detect}. However, despite modern pure artificial neural network-based models being able to perform more complex tasks alone, many AI systems continue to have ML models as components that feed into logic and other systems-of-systems to assure interpretability, performance, and even generalization in such critical applications. Many of these systems would be considered neurosymbolic (NESY) AI, a mixture of neural network methods combined with classic logic processing and programming. Defining the barriers of when a neural model becomes neurosymbolic is a topic of discussion, with some arguing modern Large Language Models (LLMs) operating in the discrete token space for inputs and outputs are neurosymbolic \cite{sarker2022neuro}.

There exists a more defined paradigm of neurosymbolic programming \cite{chaudhuri2021neurosymbolic}, which invokes neural networks as elements of a program that feed inputs to classical programs. Such programs can be written in programming languages like C and Python or declarative logic programming languages like Prolog or Datalog. These methods historically have not been able to be used with modern end-to-end deep learning paradigms, deterring adoption due to ecosystems and the ability to plug in with existing AI tools. Recent advances in differentiable reasoning have created an opportunity to incorporate reasoning and neurosymbolic programming directly into model training rather than separate post-hoc non-differentiable program additions. Having such differentiable end-to-end systems increases the applicability of many assurance testing tools \cite{ding2019advertorch} able to inspect elements and model worst-case scenarios through red-teaming for such systems.

Within this work, we begin the investigation into whether neurosymbolic programming methods can offer assurance-by-design. We do this through the creation of neurosymbolic systems and evaluating them against their fully neural counterparts across a range of tasks, modalities, and evaluations (Figure \ref{fig:example}). We note that we use the same neural backbone across all methods to better represent a fair comparison. In particular, we investigate the neurosymbolic programming language and corresponding library Scallop \cite{huang2021scallop}. Scallop has shown an advancement of neurosymbolic programming paradigms in speed, maturity, and applicability in incorporating neurosymbolic programming directly into the popular deep learning library PyTorch \cite{paszke2019pytorch} with their differentiable reasoning engine.

We perform this investigation across a handful of representational tasks and datasets. We examine multi-image logical reasoning, knowledge-base enriched image classification, visual feature reasoning for image classification, and speech recognition. We compare all methods to their fully neural counterparts across interpretability, confidence calibration, corruption robustness, adversarial robustness, and user-performance disparity. Through each dataset and task, we empirically identify gaps in current reporting on neurosymbolic reasoning's role in assurance. We find the following:

\begin{itemize}
    \item \textbf{Takeaway 1:} Differentiable neurosymbolic reasoning meets assurance needs in scenarios where there exist known logic operations, there is increasing dimensionality to inputs, and the case where neurosymbolic models vastly outperform fully neural models (Section \ref{section:results}).
    \item \textbf{Takeaway 2:} When neurosymbolic models take interpretable shortcuts, despite performance parity, the models are more vulnerable to adversarial attack (Section \ref{section:pathfinder_results}).
    \item \textbf{Takeaway 3:} The role of inference-time neurosymbolic reasoning depth has little to no effect on assurance metrics across many tasks for Scallop (Section \ref{section:roleofktesttime}).
    \item \textbf{Takeaway 4:} Differentiable neurosymbolic reasoning does not by itself offer higher assurance with less data unless the task has known logic operations where there is a large class imbalance (Section \ref{section:efficiency}).
\end{itemize}

\section{Related Work}

Prior neurosymbolic surveys \cite{agiollo2023measuring} define how neurosymbolic AI fits into definitions of trustworthiness with overlapping themes of assurance, including oversight, cross-user performance, and robustness. While there have been a plethora of claims on the trustworthiness of neurosymbolic systems \cite{wagner2024neurosymbolic}, there is a lack of a full empirical exploration of their assurance across multiple axes. Prior work has examined the promises of robust solutions in the symbolic space \cite{marconato2024not} and found that neurosymbolic systems would routinely find shortcuts in reasoning. Other works \cite{sitawarinpart} have examined the adversarial robustness of systems with deep learning models that contain intermediate symbolic representations, finding that the intermediate representation followed by another neural network was more adversarially robust. Prior work \cite{rafanelli2024empirical} has also examined the robustness of Symbolic Knowledge Injection methods to common perturbations in the input and output space for tabular data. Neurosymbolic systems have been explored for safety modeling \cite{yang2022safe} by differentiating through neurosymbolic systems for verifiable safety and assurance of worst-case scenarios. While many of these papers either mention or look at one dimension of assurance, a recent survey \cite{michel2024neuro} points out a similar lack of work outside measuring interpretability compared to other measurements of assurance. Closely related to this investigation into end-to-end neurosymbolic programming, recent work \cite{paul2024formal} examines how to expand Scallop to provide explanations towards interpretability of the answers provided by neurosymbolic methods.

\section{Background}

In this section, we cover the core concepts of differentiable neurosymbolic reasoning paradigms and the measurements of assurance relevant to these models and their fully neural counterparts.

\subsection{Differentiable Neurosymbolic Reasoning}

We address the problem where we would like to complete a task by mapping inputs, $x$, to outputs, $y$. We aim to achieve this through a learned function, $M$, combined with a symbolic program, $P$. Typically, $M$ will be learned through deep learning by parameterizing an artificial neural network with weights $\theta_M$. We assume that the outputs of $M(x)$ are then processed with a symbolic and defined program $P$, such that $P(M_\theta(x)) = y$. Generally, neurosymbolic methods separate perception of high-dimensional data to $M$ and reasoning to the program $P$. We seek to learn this function through gradient descent, optimizing $\theta_M$ to perform well given a dataset $D$ of $(x, y)$. However, complexity arises in how we define $P$ with the requirement to achieve a gradient.

Many approaches have been proposed, with the most straightforward being the application of REINFORCE \cite{williams1992simple} to optimize the outputs given samples drawn from $M_\theta$ and fed into a black-box program $P$. This provides a reward for the correct answer. However, this creates a weaker signal for the learning process and requires implementation that may not be as accessible to practitioners using common differentiable programming frameworks that perform automatic differentiation, like Torch \cite{paszke2019pytorch}. In this vein, works have implemented probabilistic reasoning domain-specific languages to both take advantage of the probability outputs and create differentiable operations in the symbolic program. We focus on methods that enable end-to-end differentiation without using reinforcement learning.

We categorize these methods as having full coverage of both the answer and potential solutions, such as DeepProbLog \cite{manhaeve2018deepproblog}; partial coverage, where the depth of potential solutions is controlled, such as Scallop; and sample-based, where samples of the answer and potential solutions are approximated. In theory, while each method has slight differences, all attempt to approximate modeling many potential outcomes given probabilities. However, as the number of outcomes and potential solutions increases, the computational cost increases dramatically. This is detrimental for learning $\theta_M$ since it typically requires many iterations over the data. Partial-coverage methods attempt to resolve this by limiting computation while still preserving the solution space. Each iteration of these methods has shown improvement, where previous ones have simply timed out \cite{huang2021scallop, solko2024data} when the complexity space increases.

Due to these factors, within this work, we examine Scallop as an intermediate method that still incorporates probabilistic reasoning while working to limit the computational intensity. While methods such as ISED \cite{solko2024data} enable more flexible program definitions through differentiating a black-box program, we seek to understand the benefits of intermediate probabilistic reasoning given the constraints of Scallop's domain-specific language.

Scallop is built on Datalog \cite{abiteboul1995foundations}, extended with probabilistic reasoning using provenance semirings. These semirings enable compositional computation on the top-k proofs, creating a differentiable function that can be used in end-to-end deep learning frameworks like PyTorch. The proofs are then calculated with success probability rather than discrete outputs for every potential proof result. An example would be adding two handwritten digits (${digit}_1 + {digit}_2$) of value 0 to 2, where the potential answers would be 0 to 4. A proof would be calculated for each of these potential answers given the potential for ${digit}_1$ and ${digit}_2$ to be 0 through 2. The value $k$ then determines for each ${digit}$ how many combinations ranked by likelihood of the joint probability between the digits would be used for the global probability of solutions. Thus at $k=1$, we would only calculate the first most likely solution. For an example and comparison of how each method in the neurosymbolic programming space works, please see the appendix of the work introducing ISED \cite{solko2024data}. For a deeper dive into the exact mechanics, we defer to the introductory paper \cite{huang2021scallop}.

\subsection{Measurements of Assurance}
\label{section:measurements_of_assurance}

Within our analysis, we examine various axes of assurance, which can at times be a loaded term. We follow the definition of the assurance of artificial intelligence systems found through rigorous analysis of requirements stated in various documents and research detailing requirements for applications of AI in critical domains \cite{batarseh2021survey}. We break down how, for this study, we measure the various aspects:

\textbf{Interpretability:} We seek to have models that can be mechanistically interpreted by developers, users, and auditors of the systems. Classically, this means we can understand why a system has made a decision. Modern deep learning models are black boxes when we treat the entire model process as a neural function. This reactionary or single-pass information processing is less interpretable. This also makes the assessment of alignment of the model problem-solving method more challenging. In contrast, when compared to treating deep learning as a function that feeds into a reasoning module, we can more easily debug, diagnose, and make informed decisions on deploying a system. For the evaluation of these methods, we rank them as more interpretable if we have some intermediate representation and trace of how the AI system came to the answer. Thus, neurosymbolic methods naturally will be more interpretable than their fully neural counterparts due to the symbolic outputs.

\textbf{Confidence Calibration:} Many models attempt to properly capture the probability of a given event using statistical modeling. Systems that can capture uncertainty can better defer to human operators or not take action in a given situation. Having high confidence calibration enables systems to be deployed with thresholds that are meaningful rather than spuriously related to misalignment with the true underlying probability. We report on our classification models with the metrics of expected calibration error (ECE) and maximum calibration error (MCE) \cite{guo2017calibration}. ECE calculates the weighted average of the absolute difference between accuracy and confidence across all bins. MCE differs in that it considers the maximum absolute difference across all the bins. Therefore, ECE is a measure of overall calibration, while MCE provides a measure of the worst-case calibration performance. These values will be between 0 and 1, with values closer to 0 indicating higher calibration.

\textbf{Corruption Robustness:} Generalization beyond the exact environment in which the model was trained is the overarching goal of ML. However, models are commonly confused by perturbations (such as an image being flipped or having parts missing). We seek to measure the performance loss given a suite of functional perturbations per task. This is the percentage loss of accuracy performance called corruption success rate (CSR), $\frac{acc - acc_{cor}}{acc}$, where $acc_{cor}$ is the accuracy under corruption.

\textbf{Adversarial Robustness:} The ability to purposefully manipulate the performance or operation of an AI system at deployment poses a significant safety risk. We define two threat models: one in which the adversary has no access during training time, so they seek to evade the correct classification using adversarial example generation with white-box access to the model (including the symbolic components). We examine a second threat in which the adversary can manipulate a percentage of training data during training in an attempt to create a backdoor trigger that can be used at deployment time. For each of these, we measure the adversary success where, at deployment time, every example is attacked with either the adversarial perturbation developed solely at deployment or the one added during training. We define adversarial success in line with \cite{papernot2016limitations} as $\frac{acc - acc_{adv}}{acc}$, resulting in a percentage of the accuracy that was degraded.

\textbf{User-Performance Parity:} Models should be performance invariant to the user's characteristics. However, ML models are commonly observed to have disparate performance \cite{zafar2017fairness}. This poses significant challenges in deploying systems that will be interacting with users, posing ethical, legal, and sensing coverage risks. Commonly, data underrepresented in the training group has lower performance than those with majority representation. We measure the accuracy disparity among users by mapping each subgroup to either the majority or minority represented meta-group. We then report the parity of accuracy through performance parity, $|acc_{maj} - acc_{min}|$, where the goal is complete parity with a value of 0.

\section{Experiments}
\label{section:experiments}

As mentioned, we use both PyTorch \cite{paszke2019pytorch} and Scallopy \cite{huang2021scallop} Python packages to define and train standard fully neural networks and neurosymbolic networks made of both neural network and symbolic reasoning components. We analyze multiple variants of Scallopy neurosymbolic models, each differing with respect to the hyperparameter \(k\), which specifies the level of reasoning granularity, allowing for a relaxation of exact probabilistic reasoning by specifying the number of top-\(k\) most likely proofs to consider.

\subsection{MNIST Logic Tasks: Multi-Image Logic Tasks}

We start with a simple series of arithmetic and logical tasks applied to the MNIST number recognition dataset. These include the \tasksum{} task and the \taskhowmany{} task. The \tasksum{} task requires prediction of the sum of all the digits represented in selected MNIST images. In contrast, the objective of the \taskhowmany{} task is to count the number of images containing the numbers 3 or 4. For each task, we defined three different variants, where the number of images drawn can be 2, 3, or 5. Each of these represents a challenging task, in particular, \tasksum{} with 5 images creates an unbalanced dataset with larger and smaller sums being less represented, which is particularly challenging for deep learning models.

We deploy two types of models for each task both with a base 3-layer residual convolutional network (CNN) \cite{he2016deep}. The first type of model serves as a baseline for a neural-only approach. This model consists of two modules. The first module comprises multiple convolutional and linear layers, embedding an image into a representation of size \(d = 1024\). Next, the embedding is fed as input to the second module, which is a two-layer multi-layer perceptron (MLP). The task of the second module is to categorize \(N\) classes in a specific task. We refer to this type of model as \textit{neural network (NN)} for the rest of the paper. The second type is the neural symbolic model in which we replace the MLP with a symbolic program that performs the operation per task. We call this type of model \textit{NESY \(k=x\)}, where \(x\) represents the number of logical reasoning steps. For our adversarial robustness evaluation, we use the \(L_\infty\)-norm with a budget of 0.03 and 100 steps to ensure convergence of the attack. For the corruption robustness, we use the corrupted datasets from MNIST-C \cite{mu2019mnist}. This totals 15 corruptions (such as fog, rotate, motion blur, and scale) with only one level of intensity.

\subsubsection{CIFAR-10: Knowledge-base Reasoning for Classification}

Knowledge bases commonly capture high-level concepts and associations that may or may not be visually grounded. Neurosymbolic methods, in particular, can incorporate these data structures to augment background information a model may not be able to pick up from data alone. We investigate the role that such knowledge bases could play by selecting the fewest number of concepts needed to disentangle all CIFAR-10 classes from one another while ensuring concept sharing. An example of such reasoning would include planes and ships sharing the concept ``is transportation'' but differing in ``has wheels''. While such background information may not always be visually grounded, such as a plane being pictured with wheels withdrawn, the background knowledge can enable models to incorporate such information in the representation.

For this reasoning task, a ResNet18 \cite{he2016deep} was used. Neural and neurosymbolic models were trained with a learning rate of \(0.001\). The fully neural model was trained to predict the 10 CIFAR classes. The neurosymbolic variant was set to predict 11 concepts that were reasoned over through the knowledge base (for details see Appendix \ref{section:cifar_logic}). The four symbolic models were trained, with \(k=1,3,5,\) and 10. For our adversarial robustness evaluation on all models, we use the \(L_\infty\)-norm with a budget of \(0.03\) and 100 steps. For the CSR, we use the CIFAR-10-C corrupted dataset \cite{hendrycks2019benchmarking} with 19 corruptions, such as Gaussian noise, motion blur, and JPEG compression, at 5 levels each.

\subsubsection{LEAF-ID: Grounded Attribute Reasoning for Classification}

Opposed to the prior experiments in image classification operating on high-level knowledge base features, the LEAF-ID dataset offers visually grounded concepts to perform intermediate reasoning. These features include shape, margin, and texture used in prior work in neurosymbolic models \cite{solko2024data}. We use a base convolutional network (CNN) model that outputs these characteristics with 3 linear heads for each concept (5 shape, 6 margin, and 4 texture concepts). This is a modified version from the prior work which used 3 CNN backbones per concept. For our fully neural model, we use the same CNN that predicts directly for the 11 classes. We train for 200 epochs, with a learning rate of \(0.0001\). We use the same equal data sampling for each of the classes pulling in a total of 40 images per class as done in prior work \cite{michel2024neuro}. With such few examples per class, this task represents the lowest data paradigm. 

We evaluate the corruption robustness by applying corruptions available from prior work \cite{hendrycks2019benchmarking}. To evaluate the adversarial robustness, we found the models to be less robust to typical strength attacks with  100\% ASR when the budget was \(0.03\) for an \(L_{\infty}\) so we decrease to where models show some level of robustness to \(0.007\) (2/255) with 200 steps. For the LEAF-ID task, we generate 19 distortions at 5 levels using the \textit{imagecorruptions} package \cite{michaelis2019dragon}, which adapts the original code from \cite{hendrycks2019benchmarking} to accommodate rectangular images.

\subsubsection{Pathfinder: Pixel-Level Reasoning}

We evaluate the robustness of operating directly in grounding to the pixel space with the Pathfinder task \cite{tay2020long}. For both models we use the base 3-layer residual CNN from the MNIST tasks. For our fully neural model, we use a linear layer on top of the convolutional layers to perform binary classification. For the symbolic model, we use two linear heads for dot and path facts. One linear head produces a dot score per pixel, with 1024 neural facts (for a 32 x 32 image). The second models the adjacency matrix where each pixel coordination that is connected to another is represented, resulting in 1,984 facts. We reduce the total number of facts from the original paper by duplicating edges in the adjacency matrix due to the connections not being directed. Due to data size and the speed of evaluating the neurosymbolic models, we limit our adversarial attacks using PGD with \(L_\infty\) to 10 steps and a budget of 0.03.

\subsubsection{Speech Word Classification: Common Voice Clips}

To further our coverage of modality beyond the commonly analyzed image modality, we also explored  assurance with speech datasets. Commonly these datasets include users of various backgrounds speaking a single word. The goal is then to train a model to perform classification to a closed-set number of discrete words. This differs from speech-to-text or transcription tasks which seek to classify multiple words in sequence. As models that perform speech transcription grow in size and compute cost, smaller models for trigger words are commonly used to save costs. This enables a lower barrier of entry for the study of non-image modalities in robustness and assurance.

Commonly, an end-to-end deep learning approach is applied directly to the speech signal to then perform a classification. However, translating the classic deep learning approach to neurosymbolic is straightforward. This is due to speech itself having a symbolic representation with the phonemic alphabet. Using phonemic elements is not a novel approach as prior state-of-the-art models, i.e. wav2vec2.0 \cite{baevski2020wav2vec} used such information bottlenecks to capture this inductive bias.

We focus on Common Voice Clips, a collection of single-word speech commands collected from users through an online platform. The classes include digits zero through nine, ``hey'', ``yes'', and ``no'', resulting in 13 classes. We remove ``Mozilla'' from the dataset due to its more unique phonemic nature and not to skew our short word modeling. We use an M5 \cite{dai2017very}, a five-layer one-dimensional CNN with a sample rate of 16 kHz and use five seconds of audio from the clip. We use a symbolic program to outline the modeling task with a maximum of six slots for phonemes. We then train a neural network to output 144 facts (max length of 6 \(\times\) 24 phonemes needed to model the space). We train both a classic neural and neurosymbolic models at a learning rate of \(0.0001\) for 250 epochs. We use Cross Entropy Loss for the neural model and Binary Cross Entropy for the neurosymbolic model. We train and test with various \(k\) values of 3, 10, 15, and 20. We perform our adversarial example generation with PGD with \(L_\infty\)-norm with a budget of \(0.001\) and 200 steps to ensure convergence. Common Voice Clips also have user labels that allow us to measure a key assurance metric of performance disparity across users. We evaluate on the English-speaking subset of the dataset. As mentioned before, the goal is to assure performance across all users independent of their training data representation.

\section{Results} \label{section:results}
In this section, we break down the results for the two MNIST logic tasks, CIFAR-10, LEAF-ID, Pathfinder, and Common Voice Clips. We also present results analyzing the role of $k$ for the NESY models during test-time with respect to assurance measurements (Section \ref{section:roleofktesttime}). Finally, we present results for how well various NN and NESY methods compare in low-data environments (Section \ref{section:efficiency}).

\subsection{MNIST Logic Tasks: Sum Results}

We see that when 2 and 3 images are input, the accuracy of all models is nearly equal, with there being little to no difference in performance (Table \ref{table:sum_performance}). Due to this high accuracy, we see that the ECE scores are nearly identical as well, with the neurosymbolic models having varied MCE scores, typically being outperformed by the neural method except for the  NESY $k=5$ case on 3 images.  For 2 and 3 images, we see a difference in the ASR between the two methods, with the NESY models having lower rates than the NN models. This is supported by the accuracy parity across models, with NN on 3 images having about 3 times the ASR as NESY. We notice here that there is a slight 0.001 increase in ASR when comparing $k=3$ and $k=5$. We later analyze the role of $k$ at test-time in Section \ref{section:roleofktesttime}.

Notably, when we increase the number of images to 5, we see a complete drop-off in performance for the NN model. This partly reflects the challenges of class imbalance mentioned earlier. The NESY models across the board outperform in every metric except MCE. We observe that due to the outputs of the NESY methods, MCE touches on an issue where the methods do not output a distribution for all combinations equal to 1, thus being penalized by this measurement even when performance is drastically higher. Even with the high accuracy, we see that the ASR for the NESY models is much lower, about 10-11\%, compared to the extremely high 99\% of the NN model given the budget. This indicates that neurosymbolic methods may offer more adversarial robustness when the combinatorics of the problem are high, there is data imbalance, and the neural architecture's learned logic operation in the MLP is increasing in dimensionality. The results for corruption robustness follow a similar trend, with corruption success being about 11\% lower for neurosymbolic methods while maintaining higher or equal accuracy. This result follows that the neurosymbolic operation is more robust than the fully learned method.

\begin{table}[h]
\centering
\begin{tabular}{c|l|r|c|c|c|c}
\Xhline{2\arrayrulewidth}
\textbf{Images} & \textbf{Model} & \textbf{Acc.} & \textbf{ECE} & \textbf{MCE} & \textbf{ASR} & \textbf{CSR} \\
\Xhline{2\arrayrulewidth}
\multirow{3}{*}{2} & NN & 0.984 & \textbf{0.012} & \textbf{0.377} & 0.080 & 0.378 \\
& NESY $k=3$ & \textbf{0.986} & \textbf{0.012} & 0.647 & \textbf{0.045} & \textbf{0.200} \\
& NESY $k=5$ & \textbf{0.986} & \textbf{0.012} & 0.483 & 0.049 & 0.263 \\
\Xhline{2\arrayrulewidth}
\multirow{3}{*}{3} & NN & 0.975 & 0.019 & 0.687 & 0.181 & 0.483 \\
& NESY $k=3$ & \textbf{0.978} & 0.019 & 0.626 & \textbf{0.066} & 0.353 \\
& NESY $k=5$ & \textbf{0.978} & \textbf{0.018} & \textbf{0.453} & 0.067 & \textbf{0.352} \\
\Xhline{2\arrayrulewidth}
\multirow{3}{*}{5} & NN & 0.283 & 0.345 & \textbf{0.599} & 0.991 & 0.584 \\
& NESY $k=3$ & 0.963 & 0.030 & 0.722 & \textbf{0.104} & 0.479 \\
& NESY $k=5$ & \textbf{0.968} & \textbf{0.028} & 0.644 & 0.113 & \textbf{0.476} \\
\Xhline{2\arrayrulewidth}
\end{tabular}
\vspace{1em}
\caption{Model performance on the \tasksum task.}
\label{table:sum_performance}
\end{table}

\subsection{MNIST Logic Tasks: How Many 3 or 4 Results}

For the results for \taskhowmany (Table \ref{table:howmany_performance}), we see parity across accuracy at all number of images. This enables a better comparison rather than the traditional neurosymbolic "can" and neural "cannot" analysis done historically in the literature and also seen in \tasksum results section. We mostly see parity across the ECE calibration metric while observing again the disadvantage that the NESY models have for MCE while typically having less calibrated scores. However, when comparing the ASR, we again see a trend that the NESY models, at all number of images and all $k$ values, have lower ASRs. We do not see a strong trend for $k$ being a determining factor for more adversarial robustness. However, there is a consistent and scaled ASR difference across tasks as the number of images increases. This again is in part because of the vulnerability of learning these symbolic operations. Again, the CSR is always lower for the NESY models while having near equal accuracy performance. We see a smaller gap (between 4-7\%) between the NN and NESY methods than CSR in the sum task until the number of images becomes 5, then we observe almost a two times decrease in performance. This again follows the idea that increasing the dimensionality of the operation space increases the potential for issues when there is a distribution shift.

\begin{table}[h]
\centering
\begin{tabular}{c|l|r|c|c|c|c}
\Xhline{2\arrayrulewidth}
\textbf{Images} & \textbf{Model} & \textbf{Acc.} & \textbf{ECE} & \textbf{MCE} & \textbf{ASR} & \textbf{CSR} \\
\Xhline{2\arrayrulewidth}
\multirow{3}{*}{2} & NN & \textbf{0.994} & \textbf{0.005} & \textbf{0.396} & 0.054 & 0.102 \\
& NESY $k=3$ & 0.993 & 0.006 & 0.576 & \textbf{0.032} & 0.067 \\
& NESY $k=5$ & 0.993 & 0.006 & 0.646 & 0.036 & \textbf{0.065} \\
\Xhline{2\arrayrulewidth}
\multirow{3}{*}{3} & NN & 0.988 & 0.011 & \textbf{0.286} & 0.100 & 0.155 \\
& NESY $k=3$ & 0.987 & 0.011 & 0.626 & \textbf{0.045} & \textbf{0.088} \\
& NESY $k=5$ & \textbf{0.989} & \textbf{0.009} & 0.719 & 0.050 & 0.101 \\
\Xhline{2\arrayrulewidth}
\multirow{3}{*}{5} & NN & 0.975 & 0.021 & \textbf{0.321} & 0.314 & 0.278 \\
& NESY $k=3$ & 0.981 & 0.017 & 0.650 & 0.090 & 0.145 \\
& NESY $k=5$ & \textbf{0.985} & \textbf{0.014} & 0.684 & \textbf{0.071} & \textbf{0.134} \\
\Xhline{2\arrayrulewidth}
\end{tabular}
\vspace{1em}
\caption{Model performance on the \taskhowmany task.}
\label{table:howmany_performance}
\end{table}

\subsection{CIFAR-10 Results}

The NN model outperformed all NESY variants on overall accuracy by approximately 6\% (see Table \ref{table:cifar_performance}). We observe that the calibration metrics tell a similar story, with NN models always better calibrated than NESY models, both by ECE and MCE metrics.  Adversarial attacks, impacted all CIFAR models similarly, with nearly 100\% ASR. While most of the ASRs are slightly lower for the NESY models, the overall accuracy under attack is still higher for the NN model. We see a similar trend with some amount of noise where the CSR is lower for NESY models, resulting in lower overall accuracy.

\begin{table}[h]
\centering
\begin{tabular}{l|c|c|c|c|c} %
\Xhline{2\arrayrulewidth}
  \textbf{Model} & \textbf{Acc.} & \textbf{ECE} & \textbf{MCE} & \textbf{ASR} & \textbf{CSR} \\
\Xhline{2\arrayrulewidth}
  NN & \textbf{0.772} & \textbf{0.127} & \textbf{0.579} & 0.972 &  \textbf{0.196} \\
  NESY k=1 & 0.715 & 0.161 & 0.664 & 0.973 & 0.186 \\
  NESY k=3 & 0.713 & 0.164 & 0.648 & \textbf{0.943} & 0.187 \\
  NESY k=5 & 0.708 & 0.171 & 0.646 & 0.964 & 0.217 \\
  NESY k=10 & 0.711 & 0.177 & 0.648 & 0.959 & 0.191 \\
\Xhline{2\arrayrulewidth}
\end{tabular} \vspace{1em}
\caption{Model performance on the CIFAR-10 task.}
\label{table:cifar_performance}
\end{table}

\subsection{LEAF-ID Results}

\begin{table}[h]
\centering
\begin{tabular}{l|c|c|c|c|c}
\Xhline{2\arrayrulewidth}
\textbf{Model} & \textbf{Acc.} & \textbf{ECE} & \textbf{MCE}  & \textbf{ASR} & \textbf{CSR} \\
\Xhline{2\arrayrulewidth}
NN  & \textbf{0.863}  & 0.066  & \textbf{0.303}  & \textbf{0.600}  & 0.302 \\
NESY k=3  & 0.818  & 0.075  & 0.367  & 0.705  & 0.297 \\
NESY k=5  & 0.850  & 0.073  & 0.498  & 0.786  & \textbf{0.275} \\
NESY k=10  & 0.809  & \textbf{0.063}  & 0.455  & 0.679 & 0.302 \\
\Xhline{2\arrayrulewidth}
\end{tabular} \vspace{1em}
\caption{Results for LEAF-ID by model.}
\label{tab:leafid_base_results}
\end{table}

For LEAF-ID, the results (Table \ref{tab:leafid_base_results}) show that there is a gap in accuracy between Neural and NESY $k=3,10$. However, NESY $k=5$ obtains close performance at 85\% relative to the NN model's 86\%. The higher k value, the better the calibration becomes, with NESY $k=10$ being slightly better in calibration through ECE. Meanwhile, we see MCE having poorer performance as $k$ increases, with a slight dropoff at $k=10$. For ASR, the NN has more robustness and accuracy under attack. There is an unclear relationship between $k$ and each attribute beyond ECE calibration. Notably, NESY $k=5$ has 61.6\% accuracy under corruptions while the NN model has 60.2\%. However, due to these small differences, we do not see large gains with having the attribute grounding.

\subsection{Pathfinder Results: Issues with Direct Pixel Grounding: Detecting Shortcuts in Symbolic System}
\label{section:pathfinder_results}

While not explored in the initial work \cite{huang2021scallop} on this dataset and task, we evaluate the interpretability of the model results. We examine how the predicted dots and connections correspond to actual dots and lines in the images. Through plotting these results, we find that models routinely and across multiple $k$ values use shortcuts in the symbolic space to accomplish the goal, similar to those reported in \cite{marconato2024not}. While the accuracy is sufficient, the models are using simple circuits in a stable region across all images. The model's behavior, visualized here, involves turning on and off a single dot to perform classifications. Thus, we find that neurosymbolic models operating in this pixel space ineffectively ground the symbols at this level. 

This behavior signals the need for additional interventions for training, beyond simply being able to plug-and-play symbols grounded to pixels. Some interventions that could be pursued include regularization-based methods to reward unique symbolic representations between examples, and architectural methods to limit the plug-and-play nature of the neurosymbolic paradigm. Mitigation for such behavior is outside the scope of this initial analysis. However, the ability to directly observe this misalignment in how the model is solving the problem is a key point in ensuring models are operating as expected from a mechanistic standpoint rather than post hoc. Despite this misalignment, we report the performance of the neurosymbolic model with an observation on the relationship between this phenomenon and adversarial robustness.

We note that the accuracy for both NN and NESY models is nearly equal (Table \ref{tab:pathfinder}). This is supported by the fact that NESY models are learning a shortcut in the neural space. The calibration metrics are also more comparable. This may be in part because this is a binary classification task. However, the main difference we see is in the ASR. We observe that the relationship between interpretability and the detection of a symbolic shortcut is closely related to poor adversarial robustness, with a general almost complete degradation in accuracy under attack, compared to the ASR of 71\% for the NN model. This would likely be reflected in the CSR as well, as the concepts are not robust to perturbation.

\begin{figure}[h]
\centering
    \includegraphics[width=\textwidth, keepaspectratio]{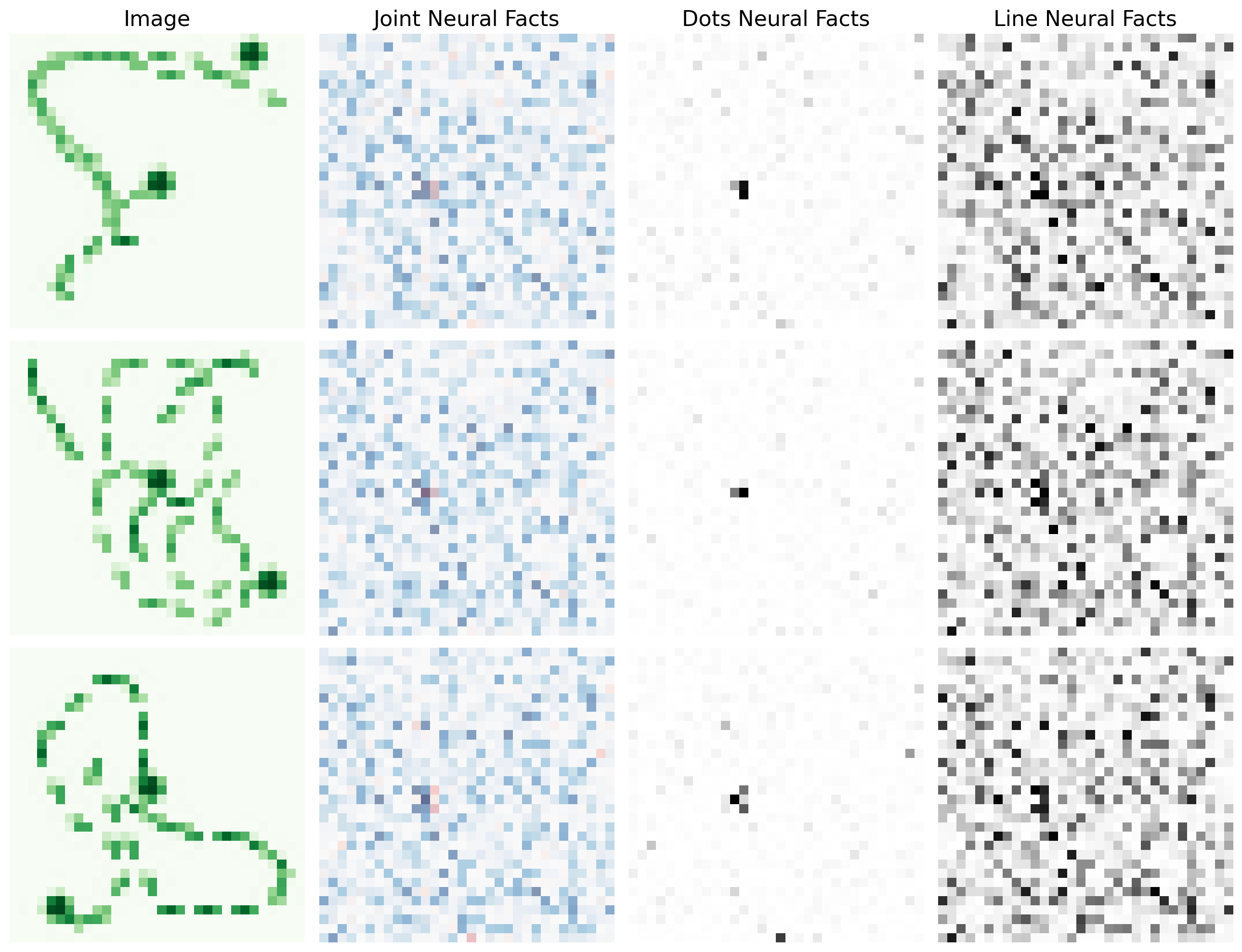}
    \caption{Internals of the Pathfinder symbolic models on the input (left), the joint neural facts imposed in the same image space to show overlap, and then separated dot and path facts with probability as the intensity of color. We see that a circuit is formed for all examples in the symbolic space that forms a shortcut for performing classification that is not grounded in actual perception.}
    \label{fig:pathfinder_1}
\end{figure}

\begin{table}[h]
\centering
\begin{tabular}{l|c|c|c|c}
\Xhline{2\arrayrulewidth}
\textbf{Model} & \textbf{Acc.} & \textbf{ECE} & \textbf{MCE}  & \textbf{ASR} \\
\Xhline{2\arrayrulewidth}
NN & 0.726 & 0.036 & 0.077 & \textbf{0.713} \\
NESY k=1 & 0.727 & \textbf{0.026} & \textbf{0.051} & 0.971 \\
NESY k=3 & \textbf{0.734} & 0.038 & 0.088 & 0.990 \\
\Xhline{2\arrayrulewidth}
\end{tabular}
\vspace{1em}
\caption{Results for Pathfinder by model.}
\label{tab:pathfinder}
\end{table}

\subsection{Common Voice Clips Results}

The results (Table \ref{tab:cv_base_results}) for speech command models indicate that while there is a slight increase in performance with respect to accuracy with $k=15$ and $20$, the calibration and adversarial success scores for symbolic models do not improve. For adversarial success, we see a consistent 20\% increase across all models. For calibration metrics, our ECE results begin to converge closer to the fully neural method as we increase the $k$ value. However, while MCE decreases as we increase $k$, it is still not close to the fully neural model.

One area where we see improvement and have a consistent trend as we increase $k$ is achieving more parity across performance for majority and minority users. This result, coupled with a small increase in accuracy, shows a lack of the common accuracy and performance trade-off commonly seen in the literature \cite{zafar2017fairness, oneto2020fairness, valdivia2021fair, kearns2019empirical}.
  
\begin{table}[h]
\centering
\begin{tabular}{l|c|c|c|c|c}
\Xhline{2\arrayrulewidth}
\textbf{Model} & \textbf{Acc.} & \textbf{ECE} & \textbf{MCE}  & \textbf{ASR} & \textbf{Dispar.} \\
\Xhline{2\arrayrulewidth}
NN  & 0.862  & \textbf{0.047}  & \textbf{0.191}  & \textbf{0.664}  & 0.065 \\
NESY k=3  & 0.848  & 0.084  & 0.688  & 0.869  & 0.112 \\
NESY k=10  & 0.855  & 0.087  & 0.466  & 0.810  & 0.075 \\
NESY k=15  & \textbf{0.877}  & 0.054  & 0.345  & 0.849  & 0.066 \\
NESY k=20  & 0.873  & 0.053  & 0.357  & 0.834  & \textbf{0.026} \\
\Xhline{2\arrayrulewidth}
\end{tabular}
\vspace{1em}
\caption{Results for Common Voice Clips by model.}
\label{tab:cv_base_results}
\end{table}

\subsection{Role of \textit{k} in Test-Time Assurance}
\label{section:roleofktesttime}

While the original Scallop work \cite{huang2021scallop} noted that they found little difference in accuracy performance, we explore how this calculation of the solution space influences other assurance metrics. In particular, we note that for adversarial attacks, including information on low-probability solutions could enable further amplification through each attack optimization step. We explore this concept across MNIST logic tasks, CIFAR-10 image classification, LEAF-ID, and Common Voice Clips audio classification.

Our results point to little to mostly no difference when applying a lower \(k\) at test time compared to train time. This is seen across both MNIST logic tasks (Table \ref{table:mnistsum_k1_results}, \ref{table:mnisthowmany_k1_results}) and CIFAR-10 (Table \ref{table:cifar_k1_results}) with differences near zero across all metrics except for a 0.1\% decrease in ASR when using only the top proof at test time. In LEAF-ID (Table \ref{tab:leafid_k1_results}), we see no difference as well. For Common Voice Clips (Table \ref{table:cv_k1_results}), we observe no to little differences in terms of the assurance results. We see a small 1\% increase in adversarial success for \(k=15\) but little to no difference for \(k=20\). We do see for \(k=20\) the model has a decrease in user disparity by 0.04, which reduces it to lower than any of the models trained. However, we do not see this for \(k=15\), which limits our ability to have a strong takeaway.

\subsection{Data Efficiency and Assurance}
\label{section:efficiency}

A major claim for neurosymbolic systems is their data efficiency. However, the majority of performance measurements so far have focused on accuracy alone. In this section, we examine how assurance measurements vary for both fully neural approaches and neurosymbolic methods. We use the same testing set with the same configurations as described in the previous section but limit the amount of training data. We hold the $k$ value stable across all models for $k=3$ for MNIST, CIFAR-10 at $k=3$, and the Common Voice Clips dataset at $k=15$. We exclude results on LEAF-ID due to the dataset already being a scarce data problem.

\subsubsection{MNIST Logic Tasks Results}

\begin{table}[h!]
    \centering
    \begin{tabular}{l|c|c|c|c|c|c}
    \hline
    & \textbf{\%} & \textbf{Acc.} & \textbf{ECE} & \textbf{MCE} & \textbf{ASR} & \textbf{CSR} \\
    \hline
    \multirow{3}{*}{NN} 
     & 5\% & 0.178 $\pm$ 0.0 & 0.528 $\pm$ 0.0 & 0.757 $\pm$ 0.01 & 0.724 $\pm$ 0.01 & 0.398 $\pm$ 0.03 \\ 
    & 10\% & 0.279 $\pm$ 0.0 & 0.447 $\pm$ 0.0 & 0.607 $\pm$ 0.01 & 0.739 $\pm$ 0.02 & 0.481 $\pm$ 0.02 \\ 
    & 25\% & 0.931 $\pm$ 0.0 & 0.040 $\pm$ 0.0 & \textbf{0.282 $\pm$ 0.02} & 0.335 $\pm$ 0.01 & 0.541 $\pm$ 0.01 \\
     \Xhline{2\arrayrulewidth}
     \multirow{3}{*}{NESY} 
     & 5\% & \textbf{0.917 $\pm$ 0.0} & \textbf{0.056 $\pm$ 0.0} & \textbf{0.430 $\pm$ 0.13} & \textbf{0.115 $\pm$ 0.0} & \textbf{0.448 $\pm$ 0.01} \\ 
    & 10\% & \textbf{0.945 $\pm$ 0.0} & \textbf{0.039 $\pm$ 0.0} & \textbf{0.427 $\pm$ 0.14} & \textbf{0.096 $\pm$ 0.0} & \textbf{0.430 $\pm$ 0.02} \\ 
    & 25\% & \textbf{0.964 $\pm$ 0.0} & \textbf{0.028 $\pm$ 0.0} & 0.419 $\pm$ 0.08 & \textbf{0.081 $\pm$ 0.0} & \textbf{0.408 $\pm$ 0.01} \\
    \hline
    \end{tabular} \vspace{1em}
    \caption{Performance on MNIST \tasksum{} of 3 images with varying levels of training data.}
    \label{table:model_performance}
\end{table}
    
The arithmetic \tasksum{} shows the most drastic difference between methods in their performance in accuracy in the small data 5\% case. The neurosymbolic method quickly reaches an accuracy in the 90\% range while the neural method requires 25\% of the data before it reaches that performance. This indicates that neurosymbolic methods perform best when implementing arithmetic or known symbolic functions. We see that the neurosymbolic method's ECE and MCE calibration metrics follow a trend of being lower due to the gap in performance. However, MCE still remains a metric where fully neural methods perform better even with lower accuracy. For the adversarial success rate (ASR), we see a drastic difference early, with the neurosymbolic method maintaining high performance across the board, with the highest value being 11\% at only 5\% of the data, while the lowest for the neural method is 35\% at 25\% of the data. This again points to the strength of having the symbolic operation versus the learned function for smaller data. An interesting trend is NESY models see more corruption robustness compared to the NN method increasing in CSR as data increases. Again, this is in the context of overall higher accuracy with NESY models on the corrupted data starting with 5\% of the data. 

\begin{table}[h!]
    \centering
    \begin{tabular}{l|c|c|c|c|c|c}
    \hline
    & \textbf{\%} & \textbf{Acc.} & \textbf{ECE} & \textbf{MCE } & \textbf{ASR} & \textbf{CSR} \\
    \hline
    \multirow{3}{*}{NN} 
    & 5\% & 0.930 $\pm$ 0.0 & 0.055 $\pm$ 0.0 & \textbf{0.254 $\pm$ 0.01} & 0.267 $\pm$ 0.01 & 0.175 $\pm$ 0.01 \\ 
    & 10\% & 0.960 $\pm$ 0.0 & 0.032 $\pm$ 0.0 & \textbf{0.259 $\pm$ 0.04} & 0.199 $\pm$ 0.02 & 0.167 $\pm$ 0.01 \\ 
    & 25\% & \textbf{0.978 $\pm$ 0.0} & \textbf{0.018 $\pm$ 0.0} & \textbf{ 0.349 $\pm$ 0.08} & 0.120 $\pm$ 0.01 & 0.163 $\pm$ 0.01 \\
    \Xhline{2\arrayrulewidth}
    \multirow{3}{*}{NESY} 
    & 5\% & \textbf{0.950 $\pm$ 0.0} & \textbf{0.035 $\pm$ 0.0} & 0.330 $\pm$ 0.13 & \textbf{0.095 $\pm$ 0.0} & \textbf{0.158 $\pm$ 0.01} \\ 
    & 10\% & \textbf{0.968 $\pm$ 0.0} & \textbf{0.024 $\pm$ 0.0} & 0.301 $\pm$ 0.05 & \textbf{0.078 $\pm$ 0.0} & \textbf{0.141 $\pm$ 0.01} \\ 
    & 25\% & \textbf{0.978 $\pm$ 0.0} & \textbf{0.018 $\pm$ 0.0} & 0.480 $\pm$ 0.26 & \textbf{0.075 $\pm$ 0.0} & \textbf{0.133 $\pm$ 0.01} \\
    \hline
    \end{tabular} \vspace{1em}
    \caption{Performance on MNIST \taskhowmany{} of 3 images with varying levels of training data.}
    \label{table:how_many_data_eff}
\end{table}

For the second MNIST logic task of \taskhowmany{}, we observe that the fully neural method has a smaller performance gap with less data, with both methods achieving above 90\% accuracy. However, we notice higher performance in ECE from the neurosymbolic approach at all percentages, indicating slightly better calibration. For MCE, the neurosymbolic method consistently performs worse, following the trend. Across all training data levels, the ASR is lower with the neurosymbolic method, remaining below 10\%, while the fully neural method is always above this threshold. The CSR across both methods is more comparable compared to the other MNIST results. The NESY models have about 2-3\% lower CSR while having higher base accuracy. For this task, both methods improve corruption robustness with more data, with NESY models having more efficiency at each step compared to NN models.  

\subsubsection{CIFAR-10 Results}

We observe that the neurosymbolic models for CIFAR-10 have lower efficiency across all assurance metrics when measuring accuracy and calibration (Table \ref{table:CIFAR-10_data_model_performance}). This starts with a near 10\% difference in accuracy and continues with NESY not able to match performance. While the ASR and CSR are at times lower, the overall accuracy performance is still higher for the neural models. For CIFAR-10 augmented with concepts, we do not observe more data efficiency along these assurance measurements.

\begin{table}[h]
\centering
\begin{tabular}{l|c|c|c|c|c|c}
\hline
 & \textbf{\%} & \textbf{Acc.} & \textbf{ECE} & \textbf{MCE} & \textbf{ASR} & \textbf{CSR} \\
\hline
\multirow{3}{*}{NN} 
    & 5\% & \textbf{0.411 $\pm$ 0.01} & \textbf{0.162 $\pm$ 0.01} & 0.590 $\pm$ 0.02 &\textbf{ 0.704 $\pm$ 0.13} & \textbf{0.094  $\pm$ 0.01} \\ 
    & 10\% & \textbf{0.501 $\pm$ 0.01} & \textbf{0.163 $\pm$ 0.0} & \textbf{0.503 $\pm$ 0.0} & \textbf{0.822 $\pm$ 0.07} & \textbf{0.131  $\pm$ 0.02} \\ 
    & 25\% & \textbf{0.588 $\pm$ 0.03} & \textbf{0.156 $\pm$ 0.0} & \textbf{0.495 $\pm$ 0.02} & 0.939 $\pm$ 0.03 & \textbf{0.187 $\pm$ 0.05} \\
\Xhline{2\arrayrulewidth}
\multirow{3}{*}{NESY} 
    & 5\% & 0.301 $\pm$ 0.02 & 0.163 $\pm$ 0.01 & \textbf{0.577 $\pm$ 0.03} & 0.680 $\pm$ 0.07 & 0.075  $\pm$ 0.0\\ 
    & 10\% & 0.415 $\pm$ 0.02 & 0.188 $\pm$ 0.01 & 0.589 $\pm$ 0.02 & 0.854 $\pm$ 0.02 & 0.104  $\pm$ 0.01\\ 
    & 25\% & 0.516 $\pm$ 0.02 & 0.203 $\pm$ 0.01 & 0.588 $\pm$ 0.02 & \textbf{0.891 $\pm$ 0.06} & 0.136  $\pm$ 0.01\\
\hline
\end{tabular} \vspace{1em}
\caption{CIFAR-10 results for varying levels of training data.}
\label{table:CIFAR-10_data_model_performance}
\end{table}

\subsubsection{Common Voice Clips Results}

Our results for Common Voice Clips (see Table \ref{table:cvclips_data_model_performance}) show nearly equal performance with small differences, but ultimately the neural models exhibit higher accuracy. For calibration metrics, we observe a larger gap, with the neural methods outperforming the neurosymbolic methods. Regarding the success rate, we find that there are small differences, with the neural model being more robust when considering both accuracy and ASR. However, we also see that disparity follows a similar trend, where the neurosymbolic model shows gains in decreasing disparity.

\begin{table}[h!]
\centering
\begin{tabular}{l|c|c|c|c|c|c}
\hline
& \textbf{\%} & \textbf{Acc.} & \textbf{ECE} & \textbf{MCE} & \textbf{ASR} & \textbf{Disparity} \\
\hline
\multirow{3}{*}{NN} 
  & 5\% & \textbf{0.446 $\pm$ 0.01} & \textbf{0.115 $\pm$ 0.01} & 0.226 $\pm$ 0.03 & 0.996 $\pm$ 0.0 & 0.050 $\pm$ 0.0 \\ 
  & 10\% & \textbf{0.579 $\pm$ 0.01} & \textbf{0.060 $\pm$ 0.0} & \textbf{0.113 $\pm$ 0.01} & \textbf{0.973 $\pm$ 0.0} & 0.070 $\pm$ 0.0 \\ 
  & 25\% & \textbf{0.756 $\pm$ 0.01} & \textbf{0.045 $\pm$ 0.01} & \textbf{0.153 $\pm$ 0.01} & \textbf{0.900 $\pm$ 0.01} & 0.024 $\pm$ 0.0 \\ 
\Xhline{2\arrayrulewidth}
\multirow{3}{*}{NESY} 
  & 5\% & 0.421 $\pm$ 0.01 & 0.162 $\pm$ 0.0 & \textbf{0.207 $\pm$ 0.01} & \textbf{0.982 $\pm$ 0.0} & \textbf{0.030 $\pm$ 0.0} \\ 
  & 10\% & 0.558 $\pm$ 0.01 & 0.156 $\pm$ 0.0 & 0.230 $\pm$ 0.01 & 0.990 $\pm$ 0.0 & \textbf{0.050 $\pm$ 0.0} \\ 
  & 25\% & 0.748 $\pm$ 0.01 & 0.153 $\pm$ 0.0 & 0.277 $\pm$ 0.01 & 0.963 $\pm$ 0.0 & \textbf{0.014 $\pm$ 0.0} \\ 
\hline
\end{tabular} \vspace{1em}
\caption{Common Voice Clips results for varying levels of training data.}
\label{table:cvclips_data_model_performance}
\end{table}

\section{Discussion}

Within our empirical analysis of both novel and existing neurosymbolic methods, we observe that there are tasks neural networks can perform and problem spaces where the assurance is evident due to performance disparity. Specifically, in scenarios where models perform equally well, only logic operations (such as counting or arithmetic) showcase better assurance metrics for neurosymbolic methods. For tasks involving categorical concept knowledge reasoning for classification, neurosymbolic models exhibit lower performance and assurance. When applying neurosymbolic methods to the symbol-to-pixel grounding problem of Pathfinder, we demonstrate that differentiable end-to-end neurosymbolic programming can take shortcuts, leading to increased adversarial vulnerability. Without the ability to inspect the model's mechanics and relying solely on accuracy, one might incorrectly choose a more performant yet shortcut-prone neurosymbolic model. However, we highlight the potential of neurosymbolic methods to create systems that perform well across different users without sacrificing overall performance when guided by symbolic programs. We also find that the depth of proof computation does not impact assurance metrics in neurosymbolic methods and that these models are not as data-efficient as fully neural methods in terms of performance and assurance unless the problem involves class imbalance.

\section{Limitations and Future Work}

While our initial empirical evaluations on the assurance of differentiable neurosymbolic reasoning showcase gaps in claims from the literature on the potential of these methods, we acknowledge that this work examined only part of the many methods emerging in mostly academic dataset contexts. The gap between well-engineered neurosymbolic systems and pure neural network methods in the real world should be examined with respect to performing proper assurance comparisons. Knowledge infusion, in particular for our tasks here, was simplistic for image classification, but greater differences could be discovered when scaling the open-endedness of visual question answering (VQA). Quantitative measurements of interpretability for detecting reasoning shortcuts, given the relation with adversarial robustness, represent a strong path forward for verification of neurosymbolic methods. Constraints and other shielding \cite{anderson2020neurosymbolic, naik2023machine} methods in the symbolic space become more relevant and ample as commonsense is more readily available for real-world use cases. The relevance of reasoning for the assurance of systems has gained attention from commercial frontier AI systems \cite{guan2024deliberative} using reasoning on safety and security policies to reject requests.

\section{Acknowledgements}

This work was conducted under the Laboratory Directed Research and Development (LDRD) Program at at Pacific Northwest National Laboratory (PNNL), a multiprogram National Laboratory operated by Battelle Memorial Institute for the U.S. Department of Energy under Contract DE-AC05-76RL01830. This article has been cleared by PNNL for public release as PNNL-SA-208413.

\bibliographystyle{plainnat}
\bibliography{main}

\appendix

\section{Appendix}

\subsection{Models Introduced}

MiniResNet is used for both the MNIST and Pathfinder tasks. The model is made up of residual blocks with channel inputs: [16, 32, 64]. 

\subsection{CIFAR-10 Model Logic}
\label{section:cifar_logic}

The concepts used are: is animal, has feathers, is mammal, has hooves, has antlers, has retractable claws, has wheels, has wings, has trailer, is transportation, and is amphibian. We use the following truth table: 

Legend for the abbreviations
\noindent
The concepts used are: \textbf{is animal (A)}, \textbf{has feathers (B)}, \textbf{is mammal (C)}, \textbf{has hooves (D)}, \textbf{has antlers (E)}, \textbf{has retractable claws (F)}, \textbf{has wheels (G)}, \textbf{has wings (H)}, \textbf{has trailer (I)}, \textbf{is transportation (J)}, and \textbf{is amphibian (K)}. 

We use the following truth table:
\begin{table}[h!]
\centering
\small
\begin{tabular}{|c|c|c|c|c|c|c|c|c|c|c|c|}
\hline
\textbf{Class} & \textbf{A} & \textbf{B} & \textbf{C} & \textbf{D} & \textbf{E} & \textbf{F} & \textbf{G} & \textbf{H} & \textbf{I} & \textbf{J} & \textbf{K} \\ \hline
Airplane & 0 & 0 & 0 & 0 & 0 & 0 & 1 & 1 & 0 & 1 & 0 \\ \hline
Automobile & 0 & 0 & 0 & 0 & 0 & 0 & 1 & 0 & 0 & 1 & 0 \\ \hline
Bird & 1 & 1 & 0 & 0 & 0 & 0 & 0 & 1 & 0 & 0 & 0 \\ \hline
Cat & 1 & 0 & 1 & 0 & 0 & 1 & 0 & 0 & 0 & 0 & 0 \\ \hline
Deer & 1 & 0 & 1 & 1 & 1 & 0 & 0 & 0 & 0 & 0 & 0 \\ \hline
Dog & 1 & 0 & 1 & 0 & 0 & 0 & 0 & 0 & 0 & 0 & 0 \\ \hline
Frog & 1 & 0 & 0 & 0 & 0 & 0 & 0 & 0 & 0 & 0 & 1 \\ \hline
Horse & 1 & 0 & 1 & 1 & 0 & 0 & 0 & 0 & 0 & 0 & 0 \\ \hline
Ship & 0 & 0 & 0 & 0 & 0 & 0 & 0 & 0 & 0 & 1 & 0 \\ \hline
Truck & 0 & 0 & 0 & 0 & 0 & 0 & 1 & 0 & 1 & 1 & 0 \\ \hline
\end{tabular} \vspace{1em}
\caption{Class Attributes Truth Table}
\label{table:cifar_logic_table}
\end{table}

\subsection{Results: Role of k at test-time}

\begin{table}[h]
\centering
\begin{tabular}{l|c|c|c|c|c}
\Xhline{2\arrayrulewidth}
\textbf{Model} & \textbf{Acc.} & \textbf{ECE} & \textbf{MCE}  & \textbf{ASR} & \textbf{CSR} \\
\Xhline{2\arrayrulewidth}
NESY k=3  & 0.977 & 0.019 & 0.625 & 0.062 & 0.352 \\
NESY k=5  & 0.975 & 0.020 & 0.698 & 0.065 & 0.393 \\
\Xhline{2\arrayrulewidth}
\end{tabular} \vspace{1em}
\caption{Results for MNIST \tasksum{} by model with k=1.}
\label{table:mnistsum_k1_results}
\end{table}

\begin{table}[h]
\centering
\begin{tabular}{l|c|c|c|c|c}
\Xhline{2\arrayrulewidth}
\textbf{Model} & \textbf{Acc.} & \textbf{ECE} & \textbf{MCE}  & \textbf{ASR} & \textbf{CSR} \\
\Xhline{2\arrayrulewidth}
NESY k=3  & 0.987 & 0.010 & 0.625 & 0.045 & 0.088 \\
NESY k=5  & 0.988 & 0.009 & 0.719 & 0.048 & 0.101 \\
\Xhline{2\arrayrulewidth}
\end{tabular} \vspace{1em}
\caption{Results for MNIST \taskhowmany{} by model with k=1.}
\label{table:mnisthowmany_k1_results}
\end{table}

\begin{table}[h!]
\centering
\begin{tabular}{l|c|c|c|c|c}
\Xhline{2\arrayrulewidth}
\textbf{Model} & \textbf{Acc.} & \textbf{ECE} & \textbf{MCE}  & \textbf{ASR} & \textbf{CSR} \\
\Xhline{2\arrayrulewidth}
NESY k=5  & 0.708 & 0.171 & 0.646 & 0.963 & 0.217 \\
NESY k=10  & 0.711 & 0.177 & 0.648 & 0.958 & 0.191 \\
\Xhline{2\arrayrulewidth}
\end{tabular} \vspace{1em}
\caption{Results for CIFAR-10 by model with k=1.}
\label{table:cifar_k1_results}
\end{table}

\begin{table}[h]
\centering
\begin{tabular}{l|c|c|c|c|c}
\Xhline{2\arrayrulewidth}
\textbf{Model} & \textbf{Acc.} & \textbf{ECE} & \textbf{MCE}  & \textbf{ASR} & \textbf{CSR} \\
\Xhline{2\arrayrulewidth}
NESY k=3  & 0.818  & 0.075  & 0.367  & 0.705  & 0.297 \\
NESY k=5  & 0.850  & 0.073  & 0.498  & 0.786  & 0.275 \\
NESY k=10  & 0.809  & 0.063  & 0.455  & 0.679 & 0.302 \\
\Xhline{2\arrayrulewidth}
\end{tabular} \vspace{1em}
\caption{Results for LEAF-ID by model.}
\label{tab:leafid_k1_results}
\end{table}

\begin{table}[h]
\centering
\begin{tabular}{l|c|c|c|c|c}
\Xhline{2\arrayrulewidth}
\textbf{Model} & \textbf{Acc.} & \textbf{ECE} & \textbf{MCE}  & \textbf{ASR} & \textbf{Disparity} \\
\Xhline{2\arrayrulewidth}
NESY k=15  & 0.877 & 0.054 & 0.345 & 0.851 & 0.066 \\
NESY k=20  & 0.873 & 0.053 & 0.357 & 0.834 & 0.022 \\
\Xhline{2\arrayrulewidth}
\end{tabular} \vspace{1em}
\caption{Results for Common Voice Clips by model with k=1.}
\label{table:cv_k1_results}
\end{table}

\end{document}